\documentclass[runningheads]{llncs}
\usepackage[T1]{fontenc}
\usepackage{graphicx}
\usepackage[dvipsnames]{xcolor}

\usepackage{multicol}
\usepackage{multirow}
\usepackage{makecell}
\usepackage{pifont}
\usepackage{setspace}

\usepackage{wrapfig}

\usepackage{amsmath,amssymb,amsfonts}
\usepackage{booktabs}
\usepackage{marvosym}
\def\ie{\textit{i.e.}}
\def\eg{\textit{e.g.}}
\def\etal{\textit{et al.}}
\def\etc{\textit{etc.}}

\usepackage[]{hyperref}
\hypersetup{
colorlinks=true,
linkcolor=red
} 
\usepackage[capitalize]{cleveref}
\crefname{section}{Sec.}{Secs.}
\Crefname{section}{Section}{Sections}
\Crefname{table}{Table}{Tables}
\crefname{table}{Tab.}{Tabs.}

\begin{document}
\title{Towards Completeness: A Generalizable Action Proposal Generator for Zero-Shot Temporal Action Localization}
%
%
\author{Jia-Run Du\inst{1}
 \and
Kun-Yu Lin\inst{1} \and
Jingke Meng\inst{1,}\textsuperscript{\Letter}\and
Wei-Shi Zheng\inst{1,2,3}
} 
\authorrunning{J.R Du et al.}
%
\institute{
    School of Computer Science and Engineering, \\
    Sun Yat-sen University, China. \\    
    \and
    Key Laboratory of Machine Intelligence and Advanced Computing, \\
    Ministry of Education, China.
    \and
    Guangdong Province Key Laboratory of Information Security Technology, \\
    Sun Yat-sen University, China. \\
    \email{mengjke@gmail.com}
}
\maketitle              
\begin{abstract}
To address the zero-shot temporal action localization (ZSTAL) task, existing works develop models that are generalizable to detect and classify actions from unseen categories. 
They typically develop a category-agnostic action detector and combine it with the Contrastive Language-Image Pre-training (CLIP) model to solve ZSTAL. 
However, these methods suffer from incomplete action proposals generated for \textit{unseen} categories, since they follow a frame-level prediction paradigm and require hand-crafted post-processing to generate action proposals.
To address this problem, in this work, we propose a novel model named Generalizable Action Proposal generator (GAP), which can interface seamlessly with CLIP and generate action proposals in a holistic way. 
Our GAP is built in a query-based architecture and trained with a proposal-level objective, enabling it to estimate proposal completeness and eliminate the hand-crafted post-processing.
Based on this architecture, we propose an Action-aware Discrimination loss to enhance the category-agnostic dynamic information of actions. 
Besides, we introduce a Static-Dynamic Rectifying module that incorporates the generalizable static information from CLIP to refine the predicted proposals, which improves proposal completeness in a generalizable manner.
Our experiments show that our GAP achieves state-of-the-art performance on two challenging ZSTAL benchmarks, \ie, Thumos14 and ActivityNet1.3. 
Specifically, our model obtains significant performance
improvement over previous works on the two benchmarks, \ie, +3.2\% and +3.4\% average mAP, respectively. The code is available at \url{https://github.com/Run542968/GAP}.

\end{abstract}

\section{Introduction}

\begin{figure}[t]
    \centering
    \includegraphics[width=1\linewidth]{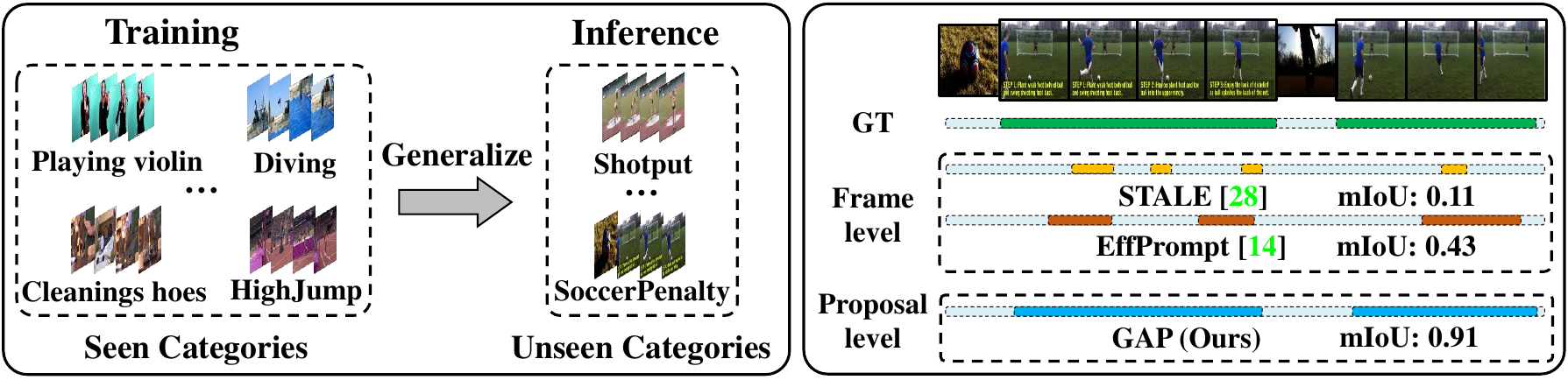}
    \vspace{-6mm}
    \caption{\textbf{Left:} Zero-shot temporal action localization requires the model trained on \textit{seen} action categories to be generalizable in detecting and classifying \textit{unseen} action categories during inference. \textbf{Right:} Visualization of the action proposals generated by STALE~\cite{nag2022zero}, EffPrompt~\cite{ju2022prompting} and our GAP. The ``mIoU'' denotes the mean Intersection over Union, which evaluates the completeness of predicted proposals. We can find that our GAP generates more complete action proposals and has a higher mIoU score than the compared frame-level methods. Best viewed in color.
    }
    \label{fig:intro}
    \vspace{-4mm}
\end{figure}

Temporal Action Localization (TAL) is one of the most fundamental tasks in video understanding, which aims to \textit{detect} and \textit{classify} action instances in long untrimmed videos. 
It is important for real-world applications such as video retrieval~\cite{lin2023univtg,wu2023cap4video,deng2023prompt,luo2023towards}, anomaly detection~\cite{sun2023hierarchical,feng2021mist,zhang2023exploiting}, action assessment~\cite{li2024continual,li2024egoexo}, and highlight detection~\cite{moon2023query,hong2020mini}. 
In recent years, many methods have shown significant performance in the close-set setting~\cite{du2022weakly,hong2021cross,liu2022end}, where categories are consistent between training and inference. 
However, a model trained in the close-set setting is capable of localizing only pre-defined action categories.
For example, a model trained on a gymnastic dataset cannot localize a ``diving'' action, even though they are both sports actions. 
As a result, temporal action localization models are significantly limited in real-world applications. 

To alleviate the above limitation, our work studies the Zero-Shot Temporal Action Localization (ZSTAL) task. This task aims to develop a localization model capable of localizing actions from \textit{unseen} categories by training with only \textit{seen} categories.
In this task, the action categories in training and inference are disjoint, that is neither labels nor data for testing categories are available during training.
For example, as shown in~\cref{fig:intro}~(Left), ZSTAL aims to develop a model that is capable of localizing instances of ``Shotput''  by training with instances of ``Diving'', ``HighJump'', \etc. 
Typically, existing works address the ZSTAL task by a composable model, which consists of a CLIP-based classifier for action classification and a category-agnostic action detector for detecting instances of unseen action categories.
For example, Ju~\etal~\cite{ju2022prompting} propose to combine the Contrastive Language-Image Pre-training (CLIP) model~\cite{radford2021learning} with an off-the-shelf frame-level action detector to solve the ZSTAL task. 
STALE~\cite{nag2022zero} design a single-stage model that consists of a parallel frame-level detector and CLIP-based classifier for ZSTAL.

Despite the progress made by these methods, they suffer from generating \textit{incomplete proposal} in detecting \textit{unseen} action categories. 
As shown in~\cref{fig:intro}~(Right), the frame-level detectors (\ie, STALE~\cite{nag2022zero} and EffPrompt~\cite{ju2022prompting}) generate fragmented action proposals and have low mIoU scores when detecting unseen category ``SoccerPenalty''.
This is because these detectors are trained with frame-level objectives and require hand-crafted post-processing (\eg, aggregating frame-level predictions via threshold) to obtain action proposals, which leads to a lack of training on estimating the completeness of action proposals.

In this work, we propose a novel Generalizable Action Proposal generator named GAP, aiming to generate complete proposals of action instances for unseen categories.
Our proposed GAP is designed with a query-based architecture, enabling it to estimate the completeness of action proposals through training with proposal-level objectives. The proposal-level paradigm eliminates the need for hand-crafted post-processing, supporting seamless integration with CLIP to address ZSTAL.
Based on the architecture, our GAP first models category-agnostic temporal dynamics and incorporates an Action-aware Discrimination loss to enhance dynamic perception by distinguishing actions from background.
Furthermore, we propose a novel Static-Dynamic Rectifying module to integrate generalizable static information from CLIP into the proposal generation process.
The Static-Dynamic Rectifying module exploits the complementary nature of static and dynamic information in actions to refine the generated proposals, improving the completeness of action proposals in a generalizable manner.
Overall, our main contributions are as follows:
\begin{itemize}
    \item
    \begin{itemize}
        We propose a novel Generalizable Action Proposal generator named GAP, which can generate action proposals in a holistic way and eliminate the complex hand-crafted post-processing.
    \end{itemize}
    \item
    \begin{itemize}
        We propose a novel Staitc-Dynamic Rectifying module, which integrates generalizable static information from CLIP to refine the generated proposals, improving the completeness of action proposals for unseen categories in a generalizable manner.
    \end{itemize}
    \item
    \begin{itemize}
        Extensive experimental results on two challenging benchmarks, \ie, Thumos14 and ActivityNet1.3, demonstrate the superiority of our method. Our approach significantly improves performance over previous work, +3.2\% and +3.4\% in terms of average mAP, on the two benchmarks, respectively.
    \end{itemize}
\end{itemize}

\section{Related Works}

\subsection{Temporal Action Localization}
Temporal Action Localization (TAL) is one of the key tasks in video understanding topics. Existing methods can be roughly divided into two categories, namely, two-stage methods and one-stage methods. The one-stage methods~\cite{zhang2022actionformer,liu2022end,shi2023tridet} do the detection and classification with a single network. Two-stage~\cite{yuan2016temporal,xu2020g,lin2021learning,lin2018bsn,lin2019bmn} methods split the localization process into two stages: proposal generation and proposal classification. Most of the previous works put emphasis on the proposal generation phrase~\cite{lin2018bsn,lin2019bmn,shi2022react,tan2021relaxed}. Concretely, boundary-based~\cite{lin2018bsn,lin2019bmn,lin2021learning} predict the probability of the action boundary and densely match the start and end timestamps according to the prediction score. Query-based methods~\cite{tan2021relaxed,shi2022react} directly generate action proposals based on
the whole feature sequence and fully leverage the global temporal context. In this work, we employ query-based architecture and focus on integrating generalizable static and dynamic information to improve the completeness of action proposals generated for \textit{unseen categories}.

\subsection{Zero-Shot Temporal Action Localization}
Zero-shot temporal action localization (ZSTAL) is concerned with the problem of detecting and classifying unseen categories that are not seen during training~\cite{nag2022zero,ju2022prompting,ju2023multi,phan2024zeetad}. This task is of significant importance for real-world applications because the available training data is often insufficient to cover all the action categories in practical use. 
Recently, EffPrompt~\cite{ju2022prompting} is the pioneering work to utilize the image-text pre-trained model CLIP~\cite{radford2021learning} for ZSTAL, which adopts an action detector (\ie, AFSD~\cite{lin2019bmn}) for action detection and apply the CLIP for action classification~\cite{lin2024human,lin2024diversifying,wang2023event,zhou2021graph,zhou2023twinformer}. Subsequently, STALE~\cite{nag2022zero} and ZEETAD~\cite{phan2024zeetad} trains a single-stage model that consists of a parallel frame-level detector and classifier for ZSTAL. Despite the process made by these methods, they struggle to generate complete action proposals for action in unseen categories.
In this work, we focus on building a proposal-level action detector, which integrates generalizable static-dynamic information to improve the completeness of action proposals.

\subsection{Vision-Language Pre-training}
The pre-trained Vision-Language Models (VLMs) have showcased significant potential in learning generic visual representation and enabled zero-shot visual recognition. As a representative work, the Contrastive Language-Image Pre-training (CLIP)~\cite{radford2021learning} was trained on 400 million image-text pairs and showed excellent zero-shot transferable ability on 30 datasets. 
In the video domain, similar ideas have also been explored for video-text pre-training~\cite{xu2021videoclip,cao2022locvtp} with a large-scale video-text dataset Howto100M~\cite{miech2019howto100m}. 
However, due to the videos containing more complex information (\eg, temporal relation) than images and large-scale paired video-text datasets being less available, video-text pre-training still has room for development~\cite{xu2021videoclip,cheng2023vindlu,huang2023clover,li2022align,lin2024rethinking,zhou2024actionhub}.
In this work, we develop a generalizable action detector that can seamlessly interface with the CLIP, thus utilizing the excellent zero-shot recognition ability of CLIP to solve the zero-shot temporal action localization problem.

\section{Methodology}

In this section, we detail our GAP, a novel Generalizable Action Proposal generator that integrates generalizable static-dynamic information to improve the completeness of generated action proposals.

\subsection{Problem Formulation} Zero-Shot Temporal Action Localization (ZSTAL) aims to \textit{detect} and \textit{classify} action instances of unseen categories in an untrimmed video, where the model is trained only with the seen categories. 
Formally, the category space of ZSTAL is divided into the seen set $C^s$ and unseen set $C^u$, where $C=C^s\cup C^u$ and $C^s\cap C^u=\varnothing$. 
Each training video $\mathcal{V}$ is labeled with a set of action annotations $\mathcal{Y}_{gt}=\{t_i,c_i\}_{i=1}^{i=N_{gt}}$, where $t_i=(t_i^s,t_i^e)$ represents the duration (\ie, action proposal) of the action instance, where $t^s_i$ and $t^e_i$ are start and end timestamps, $c_i\in C^s$ is the category and $N_{gt}$ is the number of action instances in video $\mathcal{V}$. 
In the inference phase, the model needs to predict a set of action instances $\mathcal{Y}_{pre}=\{\Tilde{t}_i,\Tilde{c}_i\}_{i=1}^{i=N_{q}}$ that has the same form as $\mathcal{Y}_{gt}$ for each video, where $N_q$ is the number of predicted action proposals in inference, and $\Tilde{c}_i\in C^u$.

\begin{figure}[t]
    \centering
    \includegraphics[width=1\linewidth]{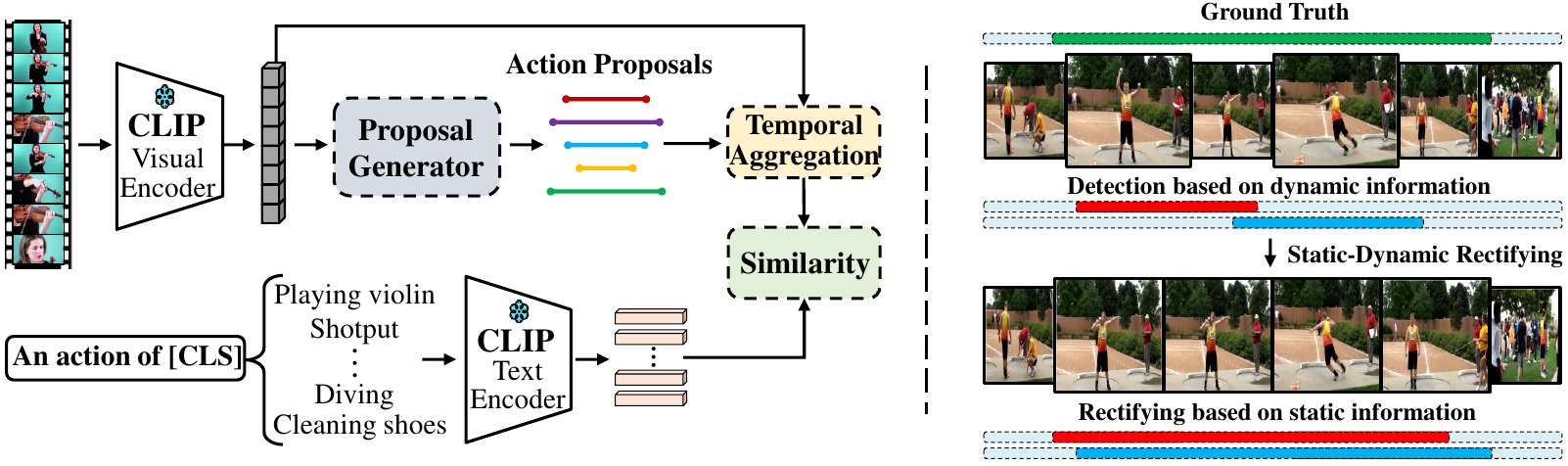}
    \vspace{-6mm}
    \caption{
        \textbf{Left:} The pipeline of our method. We adopt a video of $T=8$ with $N_q=5$ predicted action proposals for example. \textbf{Right:} An illustration of the motivation of Staitc-Dynamic Rectifying. 
        The red and blue areas in the horizontal bar represent two predicted action proposals. \textit{Top:} Detection by leveraging only dynamic information may result in incomplete proposals, where the model focuses on salient dynamic parts. \textit{Bottom:} After cooperating with static and dynamic information, the proposals are refined by interacting with proposals exhibiting consistent static information to approach ground truth. Best viewed in color.
    }
    \label{fig:framework_sdc}
    \vspace{-3mm}
\end{figure}

\begin{figure}[t]
    \centering
    \includegraphics[width=1\linewidth]{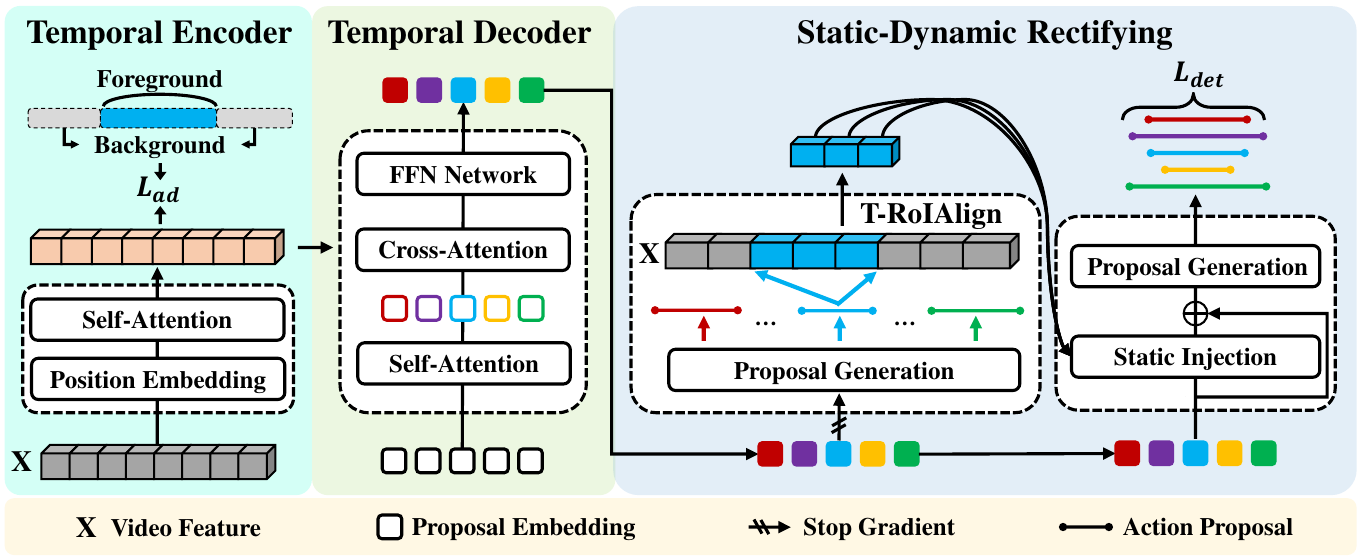}
    \caption{An illustration of our proposed GAP. Specifically, given the video feature $X$ extracted by the visual encoder, which is fed into the temporal encoder for temporal dynamics modeling. And an Action-aware Discrimination loss $\mathcal{L}_{ad}$ is used to enhance the temporal modeling by distinguishing action from the background. Next, the temporal decoder is adopted to generate dynamic-aware action queries. Then, the static information is injected into dynamic-aware action queries by the Static-Dynamic Rectifying module for refinement. Finally, action proposals are generated and supervised by the detection loss $\mathcal{L}_{det}$. Best viewed in color.
    }
    \label{fig: generator}
    \vspace{-4mm}
\end{figure}

\subsection{Model Overview}
\noindent\textbf{Pipeline of Our Method.} Our model is composed of a CLIP-based action classifier and an action detector (\ie, proposal generator), as shown in~\cref{fig:framework_sdc}~(Left).
The action detector generates category-agnostic action proposals for unseen action categories. Then, the action classification is achieved by utilizing the excellent zero-shot recognition abilities of CLIP, where a temporal aggregation module is adopted to aggregate frame features for similarity computation.

\noindent\textbf{The Proposed Action Detector.} The core of our work is the proposal-level action detector GAP, which integrates generalizable static-dynamic information to improve the completeness of generated action proposals. As shown in~\cref{fig: generator}, the GAP is designed with a query-based architecture for temporal modeling, and an Action-aware Discrimination loss $\mathcal{L}_{ad}$ is used to enhance the perception of category-agnostic temporal dynamics. Then, to mitigate the incomplete problem introduced by category-agnostic modeling, a novel Static-Dynamic Rectifying module is proposed to incorporate static information from CLIP to refine the generated proposals, improving the completeness of action proposals.

\subsection{Temporal Dynamics Modeling}
\label{sec:dynamic}
In this section, we design a query-based proposal generator with the transformer~\cite{vaswani2017attention,carion2020end} structure for temporal modeling, which incorporates an Action-aware Discrimination loss to enhance dynamics perception by distinguishing actions from background.

\vspace{-1mm}
\subsubsection{Query-based Architecture.}
Following previous works~\cite{ju2022prompting,nag2022zero}, we use the visual encoder $\mathcal{F}_{v}$ of CLIP~\cite{radford2021learning} for video feature extraction. 
Specifically, the frames of video $\mathcal{V}$ are fed into $\mathcal{F}_{v}$ to obtain features $X=\mathcal{F}_{v}(\mathcal{V}) \in \mathbb{R}^{T\times D}$, where $T$ denotes the number of frames, $D$ is the feature dimension. Subsequently, the video features $X$ are fed into the temporal encoder, where the position embedding and self-attention are applied to model the temporal relation within them. After that, the temporal features $\hat{X}\in \mathbb{R}^{T\times D}$ are obtained. 

Given the temporal features $\hat{X}$, they are fed into the temporal decoder along with a set of learnable action queries $\mathcal{Q}$. The action queries $\mathcal{Q}=\{q_i\}_{i=1}^{i=N_q}$, where $q_i$ is learnable vector with random initialization. As shown in~\cref{fig: generator}, in the decoder, the module follows the order of the self-attention module, cross-attention module, and feedforward network. Specifically, self-attention is adopted among the action queries to model the query relations with each other. The cross-attention performs the interactions between the action queries with the temporal features $\hat{X}$, thereby the action queries can integrate the rich temporal dynamics from video. Finally, the dynamic-aware action queries $\hat{\mathcal{Q}}$ are obtained after the feedforward network.

\subsubsection{Temporal Dynamics Enhancement.}
In order to enhance the temporal feature modeled by the temporal encoder, we propose an Action-aware Discrimination loss $\mathcal{L}_{ad}$ by identifying whether each frame contains an action, which is formulated  as follows:
\begin{equation}
    \mathcal{L}_{ad} = -\sum_{i=1}^{T}(m_i\log(\sigma(a_i))+(1-m_i)\log(1-\sigma(a_i))),
\end{equation}
where $\sigma$ is the sigmoid function, and $a_i$ $(i\in [1,T])$ is the actionness score for $i$-th frame, which is predicted by feeding temporal features $\hat{X}$ into a 1D convolutional network. $m_i$ is obtained by mapping the action boundary timestamps in ground truth $\mathcal{Y}_{gt}$ to temporal foreground-background mask $\{m_i\}_{i=1}^{i=T}$ as follows:
\begin{equation}
    m_i=
    \begin{cases}
      & 1,\text{ if } \frac{i}{T} \in [t^s,t^e]\\
      & 0,\text{ if } \frac{i}{T} \notin [t^s,t^e],
    \end{cases}
\end{equation}
where $[t^s,t^e]\in \mathcal{Y}_{gt}$ is the normalized [start, end] timestamps of each action instance. 

With the Action-aware Discrimination loss $\mathcal{L}_{ad}$, the temporal encoder is capable of perceiving more category-agnostic dynamics of actions, thus helping to generate more complete action proposals for unseen categories.

\subsection{Static-Dynamic Rectifying}
Since actions are composed of static and dynamic aspects~\cite{buch2022revisiting}, by only using dynamic information of action, the generator tends to predict regions exhibiting salient dynamics, rather than generating complete proposals that are close to the ground truth. For example, as shown in~\cref{fig:framework_sdc} (Right), the action proposals generated leveraging dynamic information are mainly located in the regions with intense motion in the ``Shotput'' action, such as ``turning'' and ``bending the elbow''.
Motivated by the above, we propose to integrate generalizable static and dynamic information to improve the completeness of action proposals.
We propose a Static-Dynamic Rectifying module, which injects the static information from CLIP into the dynamic-aware action queries $\hat{Q}$. As shown in~\cref{fig:framework_sdc}~(Right), by supplementing the static information, the model is aware of proposals that exhibit consistent static characteristics (\eg, contextual environment), thereby enhancing information interaction with these proposals to \textit{refine} them and improving the completeness of proposals. Notably, the Static-Dynamic Rectifying module is category-agnostic and can generalize to process unseen action categories.

Specifically, with the dynamic-aware action queries $\hat{\mathcal{Q}}$, we first feed them into the proposal generation head $\mathcal{F}_{gen}(\cdot)$ to obtain action proposals $\hat{t}=\sigma(\mathcal{F}_{gen}(\hat{\mathcal{Q}}))\in \mathbb{R}^{N_q\times 2}$, where $\sigma$ is the sigmoid function to normalize the boundary timestamps, and $\hat{t}=\{\hat{t}_i^s,\hat{t}_i^e\}_{i=1}^{i=N_q}$. Then, the static information corresponding to the action proposals is obtained by applying temporal RoIAlign~\cite{liu2022end,he2017mask} to the static feature $X$ extracted by CLIP, which is formulated as follows:
\begin{equation}
    \mathcal{Z} = \text{T-RoIAlign}(\hat{t}, X) \in \mathbb{R}^{N_q\times L \times D},
    \label{eq:roialign}
\end{equation}
where $L$ is the number of bins for RoIAlign. 
Note that the gradient back-propagation is not involved in the above process, it is only used to generate the action proposals to introduce the static information.

Subsequently, the static-dynamic action queries $\tilde{Q}$ are obtained by injecting the static features $\mathcal{Z}$ into the dynamic-aware action queries $\hat{\mathcal{Q}}$, as follows:
\begin{equation}
    \Tilde{\mathcal{Q}} = \hat{\mathcal{Q}} + \text{SA}(\text{CA}(\hat{\mathcal{Q}},\mathcal{Z})) \in \mathbb{R}^{N_q\times D}
\end{equation}
where the \textit{CA} and \textit{SA} denotes the cross-attention and self-attention, respectively. 
In this way, static information from different frames in $\mathcal{Z}$ is injected into the action query through attention-weighted aggregation.
By injecting the static information, our action queries $\Tilde{\mathcal{Q}}$ incorporate not only category-agnostic temporal dynamics from our temporal encoder but also generalizable static information from CLIP, leading to stronger cross-category detection abilities for generating complete action proposals.

\subsection{Action Proposal Generation}
\label{sec:proposal_generation}

\noindent\textbf{Proposal Generation.} Given the static-dynamic action queries $\Tilde{\mathcal{Q}}$, we feed them into the proposal generation head $\mathcal{F}_{gen}(\cdot)$ to generate category-agnostic action proposals $\Tilde{t}=\sigma(\mathcal{F}_{gen}(\Tilde{\mathcal{Q}}))\in \mathbb{R}^{N_q\times 2}$, where $\sigma$ is the sigmoid function to normalize the boundary timestamps, and $\Tilde{t}=\{\Tilde{t}_i^s,\Tilde{t}_i^e\}_{i=1}^{i=N_q}$. 

In addition, along with generated action proposals, we predict category-agnostic foreground probabilities $\mathcal{E} = \sigma(\mathcal{F}_{cls}(\Tilde{\mathcal{Q}})) \in \mathbb{R}^{N_q}$ for action proposals, where $\mathcal{F}_{cls}$ is the binary classification head and $\mathcal{E}=\{\xi_i\}_{i=1}^{i=N_q}$.

\noindent\textbf{Category-Agnostic Detection Loss.} Given the action proposals $\Tilde{t}$, their foreground probabilities $\mathcal{E}$ and the ground-truth action proposals $t=\{t_i^s,t_i^e\}_{i=1}^{i=N_{gt}}$.
Similar to DETR~\cite{carion2020end}, we assume $N_q$ is larger than $N_{gt}$ and the ground-truth action proposals $t$ is augmented to be size $N_q$ by padding $\varnothing$. Then, the category-agnostic detection loss $\mathcal{L}_{det}$ is given as follows:
\begin{equation}
    \mathcal{L}_{det} =  \sum_{j=1}^{N_q}[\mathcal{L}_{cls}(\xi_{\hat{\pi}(j)},\xi^*)+\mathbb{I}_{t_j\ne \varnothing} \mathcal{L}_{reg}(\Tilde{t}_{\hat{\pi}(j)},t_j)],
    \label{eq:det_loss}
 \end{equation}
where $\mathcal{L}_{reg}=\mathcal{L}_{1} + \mathcal{L}_{tIoU}$, and $\mathcal{L}_{cls}$ is the binary classification loss that is implemented via focal loss~\cite{lin2017focal}. $\xi^*$ is $1$ if the sample is marked positive, and otherwise $0$. The $\hat{\pi}$ is the permutation that assigns each ground truth to the corresponding prediction, it is obtained by Hungarian algorithm~\cite{kuhn1955hungarian} as follows:
\begin{equation}
    \hat{\pi} = \arg \min\sum_{i=1}^{N_q}Cost(\Tilde{t}_i,\xi_i,t_i),
    \label{eq:hungarian}
\end{equation}
where $Cost(\Tilde{t}_i,\xi_i,t_i)$ is defined as $\mathbb{I}_{\{t_i\ne \varnothing\}} [\alpha\cdot \mathcal{L}_1(\Tilde{t}_i,t_i) - \beta\cdot \mathcal{L}_{tIoU}(\Tilde{t}_i,t_{i}) - \gamma\cdot \xi_i]$, and $\mathcal{L}_{tIoU}$ is the temporal IoU loss~\cite{liu2022end}

\subsection{Training Objective and Inference}

\noindent\textbf{Training Objective.} Overall, the training objective of our GAP is given as follows:
\begin{equation}
    \mathcal{L} = \mathcal{L}_{det} + \lambda_{ad}\cdot\mathcal{L}_{ad},
\end{equation}
where $\lambda_{ad}=3$ and the balance factor of $\mathcal{L}_{cls}$, $\mathcal{L}_{1}$ and $\mathcal{L}_{tIoU}$ in $\mathcal{L}_{det}$ are $3$, $5$ and $2$, respectively.

\noindent\textbf{Zero-Shot Inference.}
After generating the category-agnostic action proposals, following previous works~\cite{nag2022zero,ju2022prompting}, we construct the text prompt to transfer the zero-shot recognition capability of CLIP, as shown in~\cref{fig:framework_sdc} (Left). 

Specifically, the category name is wrapped in a prompt template ``\textit{a video of a person doing} $<CLS>$'', then the textual (\ie, prompt) embeddings $\mathcal{S}\in \mathbb{R}^{N_c\times D}$ are obtained by feeding the prompt into text encoder $\mathcal{F}_{t}$ of CLIP, where  $N_c$ is the number of \textit{unseen} categories.

Given the category-agnostic action proposals $\Tilde{t}$ generated by the action detector, we obtain the frame features $\mathcal{Z}\in \mathbb{R}^{N_q\times L\times D}$ corresponding to action proposals by applying the temporal RoIAlign to spatial features $X$, as in~\cref{eq:roialign}. Subsequently, the action classification is conducted as follows:
\begin{equation}
    \hat{c} = \mathop{\arg\max}\limits_{c\in N_c} \psi (\cos(\mathcal{Z},\mathcal{S})) \in \mathbb{R}^{N_q},
\end{equation}
where $\hat{c}=\{\Tilde{c}_i\}_{i=1}^{i=N_q}$ is the set of predicted categories corresponding to the action proposals, $\psi$ is the temporal aggregation module and $cos(\cdot,\cdot)$ denotes the cosine similarity. Subsequently, the final prediction $\mathcal{Y}_{pre}=\{\Tilde{t}_i,\Tilde{c}_i\}_{i=1}^{i=N_q}$ is obtained by combining the predicted action proposals $\Tilde{t}$ and predicted category $\hat{c}$.

\section{Experiments}
\subsection{Datasets and Evaluation Metrics}
We evaluate our method on two public benchmarks, \ie, Thumos14~\cite{THUMOS14} and ActivityNet1.3~\cite{caba2015activitynet}, for zero-shot temporal action localization. Following the previous methods~\cite{ju2022prompting,nag2022zero}, we adopt two split settings for zero-shot scenarios: (1) training with 75\% action categories and test on the left 25\% action categories; (2) training with 50\% categories and test on the left 50\% action categories.

\noindent\textbf{Thumos14} contains 200 validation videos and 213 test videos of 20 action classes. It is a challenging benchmark with around 15.5 action instances per video and whose videos have diverse durations. We use the validation videos for training and the test videos for test, following previous works.

\noindent\textbf{ActivityNet1.3} is a large dataset that covers 200 action categories, with a training set of 10,024 videos and a validation set of 4,926 videos. It contains around 1.5 action instances per video. We use the training and validation sets for training and test, respectively. 

\noindent\textbf{Evaluation metric}. Following previous works~\cite{nag2022zero,ju2022prompting}, we evaluate our method by mean average precision (mAP) under multiple IoU thresholds, which are standard evaluation metrics for temporal action localization. Our evaluation is conducted using the officially released evaluation code~\cite{caba2015activitynet}.
Moreover, to evaluate the quality of proposals generated by our method, we calculate Average Recall (AR) with Average Number (AN) of proposals and area under AR \textit{v.s.} AN curve per video, which are denoted by AR@AN and AUC. Following the standard protocol~\cite{lin2019bmn}, we use tIoU thresholds set [0.5:0.05:1.0] on Thumos14 and [0.5:0.05:0.95] on ActivityNet1.3 to calculate AR@AN and AUC.

\subsection{Implementation Detatils}
\label{sec:details}
For a fair comparison with previous works~\cite{ju2022prompting,nag2022zero}, we \textit{only} adopt the visual and text encoders from pre-trained CLIP~\cite{radford2021learning} (ViT-B/16) to extract video and text prompt features, the dimension $D=512$. 
The number of layers for the temporal encoder and decoder for Thumos14 and ActivityNet1.3 is set to 2, 5, and 2, 2 respectively. The proposal generation head, binary classification head, and temporal aggregation module are implemented by MLP, FC, and average pooling, respectively. 
The AdamW~\cite{loshchilov2017decoupled} optimizer with the batch size $16$ and weight decay $1\times 10^{-4}$ is used for optimization. 
The equilibrium coefficients $\alpha$, $\beta$ and $\gamma$ in~\cref{eq:hungarian} are specified as $5$, $2$ and $2$. The number of bins $L=16$ for RoIAlign. The number of action queries is set to $40$ and $30$, learning rate is set to $1\times 10^{-4}$ and $5\times 10^{-5}$ for Thumos14 and ActivityNet1.3. The method is implemented in PyTorch~\cite{paszke2019pytorch} and all experiments are performed on an NVIDIA GTX 1080Ti GPU. More details are available in \textit{supplementary material}.

\subsection{Comparison with State-of-the-Arts}
\noindent\textbf{Performance of localization results.} In~\cref{tab:sota}, we compare our method with the state-of-the-art ZSTAL methods on Thumos14 and ActivityNet1.3 datasets, in terms of mAP metric.
From the results, it can be found that our method significantly outperforms the existing methods and achieves new state-of-the-art performance on both datasets. 
Our method outperforms the latest method by $3.2$\% and $3.4$\% in terms of average mAP (\ie, AVG) of the 75\% \textit{v.s.}~25\% split on the Thumos14 and ActivityNet1.3 datasets, respectively. 
In the case of the more challenging 50\% \textit{v.s.}~50\% split, our method still significantly outperforms the state-of-the-art methods.
This demonstrates the effectiveness of our proposed proposal-level action detector. 
It is worth noting that for a fair comparison with other methods, we only use CLIP (\ie, RGB only) as the backbone, without the introduction of optical flow features that necessitate complex processing.
This demonstrates that our GAP has excellent generalization ability to detect the location of unseen action categories by integrating generalizable static and dynamic information.

\noindent\textbf{Quality of generated action proposals.} We conduct a comparison between our proposed GAP and existing methods in terms of the quality of generated action proposals for unseen action categories. All experiments are performed in the split 75\% \textit{v.s.} 25\% on the Thumos14 dataset. Notably, the ZEETAD~\cite{phan2024zeetad} does not release its code, so we cannot make a fair comparison with it.
Following the standard protocol~\cite{lin2019bmn}, we adopt the AR@AN and AUC as evaluation metrics, and the comparison results are summarized in table~\cref{tab:compare_ar}. 
From the results, we can find that our method significantly outperforms the previous ones in both AR and AUC metrics. This demonstrates that our GAP can generate more accurate and complete action proposals for unseen actions. 
This is attributed to both the proposed proposal-level detector and the integration of generalizable static and dynamic information, which significantly improves the generalizability to detect actions from unseen categories.

\begin{table}[t] 
	\centering
     \centering\caption{Comparison with the state-of-the-art ZSTAL methods on Thumos14 and ActivityNet1.3 datasets. AVG represents the average mAP (\%) computed under different IoU thresholds, \ie, [0.3:0.1:0.7] for Thumos14 and [0.5:0.05:0.95] for ActivityNet1.3. The $\dagger$ denotes the extra information (\ie, optical flow) is disabled for a fair comparison. All results of the compared methods are from their official report.}
    \scalebox{0.8}{
		\begin{tabular}{cccccccccccc@{\hspace{0.2em}}c}
            \toprule[1pt]
		    \multirow{2}{*}{\textbf{Split}} & \multirow{2}{*}{\textbf{Method}} & \multicolumn{7}{c}{\textbf{Thumos14}} & \multicolumn{4}{c}{\textbf{ActivityNet1.3}} \\

		    \cmidrule{3-8} \cmidrule{10-13}
		    & & \textbf{0.3} & \textbf{0.4} & \textbf{0.5} & \textbf{0.6} & \textbf{0.7} & \textbf{AVG} & & \textbf{0.5} & \textbf{0.75} & \textbf{0.95} & \textbf{AVG} \\ 
		    \midrule
            \multirow{5}{*}{\makecell[c]{75\% Seen \\ 25\% Unseen}}
            & DenseCLIP~\cite{rao2022denseclip} & 28.5 & 20.3 & 17.1 & 10.5 & 6.9 & 16.6 & &32.6 &18.5 &5.8 &19.6\\
            & CLIP~\cite{radford2021learning} & 33.0 & 25.5 & 18.3 & 11.6 & 5.7 & 18.8 & &35.6 &20.4 &2.1 &20.2\\
            & EffPrompt~\cite{ju2022prompting} & 39.7 & 31.6& 23.0 & 14.9 & 7.5 & 23.3 & &37.6 &22.9 &3.8 &23.1\\ 
            & STALE~\cite{nag2022zero} & 40.5 & 32.3 & 23.5 & 15.3 & 7.6 & 23.8 & &38.2 &25.2 &6.0 &24.9\\
            & ZEETAD~\cite{phan2024zeetad}$^\dagger$ & 47.3 & - & 29.7 & - & 11.5 & 29.7 & &45.5&28.2&6.3 &28.4\\
    
            \cmidrule{2-13}
            & \textbf{Ours} & \textbf{52.3} & \textbf{44.2} & \textbf{32.8} & \textbf{22.4} & \textbf{12.6} & \textbf{32.9} & & \textbf{47.6} & \textbf{32.5} & \textbf{8.6} & \textbf{31.8}\\ 
            \midrule

            \multirow{5}{*}{\makecell[c]{50\% Seen \\ 50\% Unseen}}
            & DenseCLIP~\cite{rao2022denseclip} & 21.0 & 16.4 & 11.2 & 6.3 & 3.2 & 11.6 & &25.3 &13.0 &3.7& 12.9\\
            & CLIP~\cite{radford2021learning} & 27.2 & 21.3 & 15.3 & 9.7 & 4.8 & 15.7& &28.0& 16.4 &1.2 &16.0 \\
            & EffPrompt~\cite{ju2022prompting} & 37.2 & 29.6 & 21.6 & 14.0 & 7.2 & 21.9 &&32.0 &19.3 &2.9 &19.6\\ 
            & STALE~\cite{nag2022zero} & 38.3 & 30.7 & 21.2 & 13.8&  7.0 & 22.2& &32.1 &20.7 &5.9 &20.5\\
            
            \cmidrule{2-13}
            & \textbf{Ours} & \textbf{44.2} & \textbf{36.0} & \textbf{27.1} & \textbf{15.1} & \textbf{8.0} & \textbf{26.1} &  & \textbf{41.6} &\textbf{26.2} & \textbf{6.1} & \textbf{26.4}\\ 

            \bottomrule[1pt]
		\end{tabular}
		}
      \label{tab:sota}
\end{table}

\begin{table}[t] 
	\centering
    \fontsize{8.5pt}{11.5pt}\selectfont
     \centering\caption{Ablation studies of our method on the Thumos14 dataset, adopting the 75\% \textit{v.s.}~25\% split. The ``Actionness'' denotes the Action-aware Discrimination loss $\mathcal{L}_{ad}$ and ``Rectifying'' denotes the Static-Dynamic Rectifying module.}
    \vspace{-2mm}
    \scalebox{1}{
		\begin{tabular}{lccccccccccc}
		    \toprule[1pt]
		    \multirow{2}{*}{\textbf{Models}} & \multicolumn{6}{c}{\textbf{mAP@IoU}} & \multicolumn{4}{c}{\textbf{AR@AN}} & \multirow{2}{*}{\textbf{AUC}} \\
            \cmidrule{2-7} \cmidrule{9-11} 
             & \textbf{0.3} & \textbf{0.4} & \textbf{0.5} & \textbf{0.6} & \textbf{0.7} & \textbf{AVG}  && \textbf{@10} & \textbf{@25} & \textbf{@40} &   \\
		    \midrule

            Full & \textbf{52.3} & \textbf{44.2} &\textbf{32.8}& \textbf{22.4} & \textbf{12.6} &\textbf{32.9} && \textbf{12.7}  &\textbf{22.7} & \textbf{25.6} &\textbf{23.8} \\
            w/o Rectifying & 50.6 &39.7 & 31.8 & 19.8& 10.5 &30.5 && 12.3  & 21.1 & 23.9 &22.6 \\
            w/o Rectifying \& Actionness & 49.0 &39.7& 28.7&17.7& 8.2 &28.7 && 11.4 & 20.5  & 22.9 &21.6 \\
            \bottomrule[1pt]
		\end{tabular}}
      \label{tab:ablation}
\end{table}

\begin{table}[htbp]
    \begin{minipage}{0.49\textwidth}
      \centering
       \centering\caption{Comparison with the state-of-the-art ZSTAL methods in terms of AR@AN (\%) and AUC (\%). ``Frame'' and ``Proposal'' denote the frame-level and the proposal-level detector, respectively. 
       }
      \scalebox{0.8}{
          \begin{tabular}{l@{\hspace{0.25em}}ccccc}
              \toprule[1pt]
              \multirow{2}{*}{\textbf{Method}}&\multirow{2}{*}{\textbf{\makecell[c]{Detector\\ Type}}} & \multicolumn{3}{c}{\textbf{AR@AN}} & \multirow{2}{*}{\textbf{AUC}} \\
              \cmidrule{3-5}
              & & \textbf{@10} & \textbf{@25} & \textbf{@40} & \\
              \midrule
              EffPrompt~\cite{ju2022prompting} &Frame&9.3&15.7&19.6&19.3\\
              STALE~\cite{nag2022zero} &Frame&6.9&12.6&15.8&14.8\\
              \midrule
              Ours &Proposal&\textbf{12.7}&\textbf{22.7}&\textbf{25.6}&\textbf{23.8}\\
              \bottomrule[1pt]
          \end{tabular}
          }
        \label{tab:compare_ar}
    \end{minipage}%
    \hspace{2mm}
    \begin{minipage}{0.49\textwidth}
      \centering
       \centering\caption{Comparison of different implementations of Static-Dynamic Rectifying module. All experiments are performed in the split 75\% \textit{v.s.} 25\% on Thumos14.}
      \scalebox{0.8}{
          \begin{tabular}{lccccc}
              \toprule[1pt]
              \multirow{2}{*}{\textbf{Models}}&\multirow{2}{*}{\textbf{AVG}} & \multicolumn{3}{c}{\textbf{AR@AN}} & \multirow{2}{*}{\textbf{AUC}} \\
              \cmidrule{3-5}
              & & \textbf{@10} & \textbf{@25} & \textbf{@40} & \\
              \midrule
              STALE~\cite{nag2022zero} &23.8&6.9&12.6&15.8&14.8\\
              \midrule

              Mean &30.3&12.0&22.1&24.6&23.2\\
              Max &31.8&12.5&21.8&24.7&23.4\\
              Cross-Attention &\textbf{32.9}&\textbf{12.7}&\textbf{22.7}&\textbf{25.6}&\textbf{23.8}\\
              \bottomrule[1pt]
          \end{tabular}
          }
        \label{tab:different_impl}
    \end{minipage}
    \vspace{-6mm}
  \end{table}

\subsection{Analysis}
We conduct extensive quantitative and qualitative analysis to demonstrate the effectiveness of our proposed GAP. All experiments are performed in the split 75\% \textit{v.s.} 25\% on the Thumos14 dataset. More analyses are available in \textit{supplementary material}.

\noindent\textbf{Ablation studies of each component.} In~\cref{tab:ablation}, we show the quantitative analysis of the different components in our method. 
By comparing the first and second rows, removing the Static-Dynamic Rectifying module results in the $2.4$\% and $1.2$\% performance degradation in terms of AVG and AUC, which demonstrates that the integration of generalizable static-dynamic information does help to improve the detection abilities of the detector to generalize to unseen action categories. 
From the second and third rows, we find that the absence of the Action-aware Discrimination loss $\mathcal{L}_{ad}$ leads to a $1.8$\% and $1.0$\% performance drop of AVG and AUC, respectively. 
This is attributed to that $\mathcal{L}_{ad}$ enhances the ability of the temporal encoder to perceive category-agnostic dynamic information.
Moreover, from the third row, we find that by only adopting the category-agnostic detector, our method still outperforms the frame-level method STALE~\cite{nag2022zero} $4.9$\% and $6.8$\% in terms of AVG and AUC.
This is because the frame-level detector in STALE generates action proposals by grouping consecutive frames, resulting in fragmented action proposals. Our proposed proposal-level detector is able to generate action proposals directly, which guarantees the completeness of action proposals in a holistic way.

\noindent\textbf{Different implementations of Static-Dynamic Rectifying module.} In~\Cref{tab:different_impl}, we compare the different implementations of the Static-Dynamic Rectifying module. 
``Mean'' and ``Max'' refer to the static information of different frames (\ie, $L$) in $\mathcal{Z}\in \mathbb{R}^{N_q\times L\times D}$ aggregated through average pooling and max pooling, respectively. 
From the results, we find that the best performance is achieved by adopting cross-attention, which is attributed to the attention-adaptive aggregation focusing on more valuable information.
Notably, regardless of different implementations, our method still outperforms the state-of-the-art method STALE~\cite{nag2022zero} in all metrics. This demonstrates that combining generalizable static-dynamic information effectively improves the generalization ability of our GAP to detect unseen action categories.

\begin{figure}[t]
    \begin{minipage}{0.49\textwidth}
        \centering
        \includegraphics[width=\textwidth]{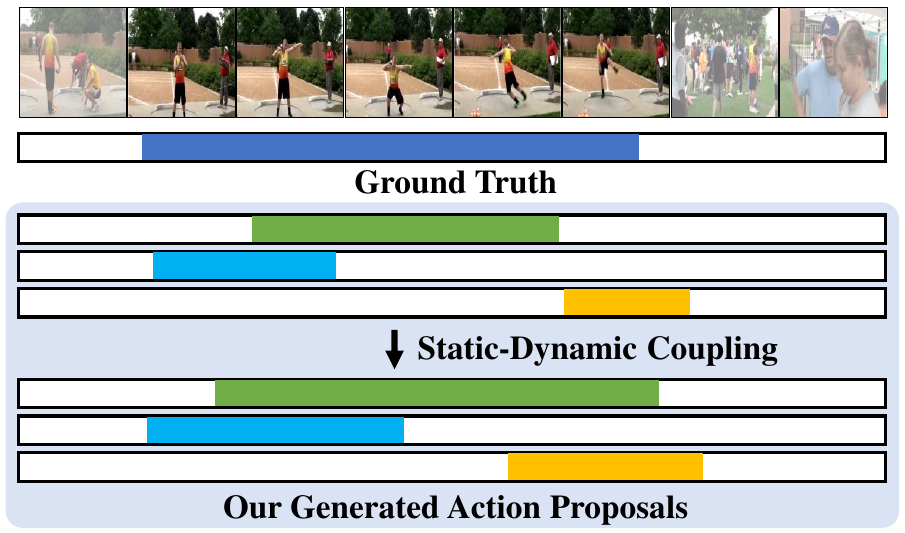}
        \caption{Visualization of the three action proposals before and after the Static-Dynamic Rectifying module, without retraining. The same color represents the result from the same action proposal. Best viewed in color.}
        \label{fig:sdc_visual}
    \end{minipage}%
    \hspace{2mm}
    \begin{minipage}{0.49\textwidth}
        \centering
        \includegraphics[width=1\linewidth]{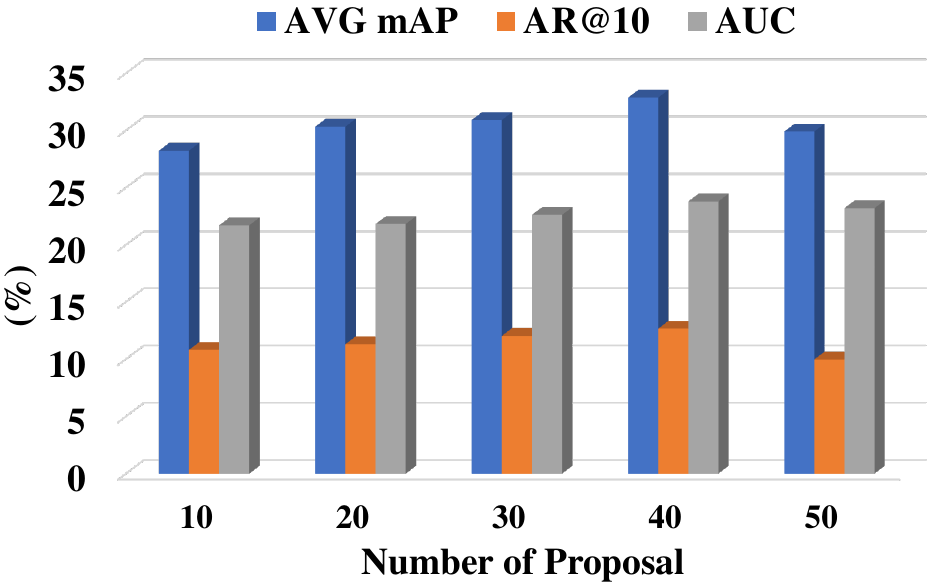}
        \vspace{-8mm}
        \caption{Performance of different number of action queries. AVG mAP denotes the average mAP for IoU thersholds from 0.1 to 0.7 with 0.1 increment. All experiments are performed in the split 75\% \textit{v.s.} 25\% on the Thumos14 dataset. Best viewed in color.
        }
        \label{fig:num_queries}
    \end{minipage}
    \vspace{-5mm}
\end{figure}

\noindent\textbf{Qualitative analysis of Static-Dynamic Rectifying.} In~\cref{fig:sdc_visual}, we track and visualize the changes in the specified action proposals before and after applying the Static-Dynamic Rectifying module. 
Note that here the \textit{input} and \textit{output} of the Static-Dynamic Rectifying module are compared directly, without retraining. 
The experiments are performed on our full method, and we choose the top-$3$ category-agnostic action proposals with the highest predicted scores for visualization. 
From the result, we find that the durations (start, end) of the three different action proposals are all refined after the Static-Dynamic Rectifying module. This further verifies that the Static-Dynamic Rectifying module improves the completeness of action proposals by exploiting the complementary nature of static-dynamic information.

\noindent\textbf{Analysis of the number of action queries.} In~\Cref{fig:num_queries}, we compare the results under different number of action queries. Due to the query-based architecture we adopted, each action query in our action detector corresponds to an action proposal. In principle, a fewer number of action queries results in missing action instances of unseen categories, while a large number of action queries results in generating a large number of low-quality action proposals. As shown in~\Cref{fig:num_queries}, our method achieves the best performance when using a medium number of action queries (\ie, 40 queries). Despite the varied performance using different numbers of action queries, our proposed GAP can outperform state-of-the-arts in all the cases as shown in the figure, which demonstrates our effectiveness in generating high-quality action proposals.

\section{Conclusion}
We propose a novel Generalizable Action Proposal generator named GAP, which can generate more complete action proposals for unseen action categories compared with previous works. Our GAP is designed with a query-based architecture, enabling it to generate action proposals in a holistic way. The GAP eliminates the need for hand-crafted post-processing, supporting seamless integration with CLIP to solve ZSTAL. Furthermore, we propose a novel Staitc-Dynamic Rectifying module, which integrates generalizable static and dynamic information to improve the completeness of action proposals for unseen categories. Extensive experiments on two datasets demonstrate the effectiveness of our method, and our approach significantly outperforms previous methods, achieving a new state-of-the-art performance.

\subsubsection{Acknowledgements} This work was supported partially by NSFC (No.62206315), Guangdong NSF Project (No.2023B1515040025, No.2024A1515010101), Guangzhou Basic and Applied Basic Research Scheme (No.2024A04J4067).

%
%
%
{
\bibliographystyle{splncs04}
\bibliography{main}

\begin{thebibliography}{10}
\providecommand{\url}[1]{\texttt{#1}}
\providecommand{\urlprefix}{URL }
\providecommand{\doi}[1]{https://doi.org/#1}

\bibitem{buch2022revisiting}
Buch, S., Eyzaguirre, C., Gaidon, A., Wu, J., Fei-Fei, L., Niebles, J.C.: Revisiting the" video" in video-language understanding. In: CVPR (2022)

\bibitem{caba2015activitynet}
Caba~Heilbron, F., Escorcia, V., Ghanem, B., Carlos~Niebles, J.: {ActivityNet: A Large-Scale Video Benchmark for Human Activity Understanding}. In: CVPR (2015)

\bibitem{cao2022locvtp}
Cao, M., Yang, T., Weng, J., Zhang, C., Wang, J., Zou, Y.: Locvtp: Video-text pre-training for temporal localization. In: ECCV (2022)

\bibitem{carion2020end}
Carion, N., Massa, F., Synnaeve, G., Usunier, N., Kirillov, A., Zagoruyko, S.: End-to-end object detection with transformers. In: ECCV (2020)

\bibitem{cheng2023vindlu}
Cheng, F., Wang, X., Lei, J., Crandall, D., Bansal, M., Bertasius, G.: Vindlu: A recipe for effective video-and-language pretraining. In: CVPR (2023)

\bibitem{deng2023prompt}
Deng, C., Chen, Q., Qin, P., Chen, D., Wu, Q.: Prompt switch: Efficient clip adaptation for text-video retrieval. In: ICCV (2023)

\bibitem{du2022weakly}
Du, J.R., Feng, J.C., Lin, K.Y., Hong, F.T., Wu, X.M., Qi, Z., Shan, Y., Zheng, W.S.: Weakly-supervised temporal action localization by progressive complementary learning. arXiv  (2022)

\bibitem{feng2021mist}
Feng, J.C., Hong, F.T., Zheng, W.S.: Mist: Multiple instance self-training framework for video anomaly detection. In: CVPR (2021)

\bibitem{he2017mask}
He, K., Gkioxari, G., Doll{\'a}r, P., Girshick, R.: Mask r-cnn. In: ICCV (2017)

\bibitem{hong2021cross}
Hong, F.T., Feng, J.C., Xu, D., Shan, Y., Zheng, W.S.: {Cross-modal Consensus Network for Weakly Supervised Temporal Action Localization}. In: ACM MM (2021)

\bibitem{hong2020mini}
Hong, F.T., Huang, X., Li, W.H., Zheng, W.S.: Mini-net: Multiple instance ranking network for video highlight detection. In: ECCV (2020)

\bibitem{huang2023clover}
Huang, J., Li, Y., Feng, J., Wu, X., Sun, X., Ji, R.: Clover: Towards a unified video-language alignment and fusion model. In: CVPR (2023)

\bibitem{THUMOS14}
Jiang, Y.G., Liu, J., Roshan~Zamir, A., Toderici, G., Laptev, I., Shah, M., Sukthankar, R.: {THUMOS} challenge: Action recognition with a large number of classes. \url{http://crcv.ucf.edu/THUMOS14/} (2014)

\bibitem{ju2022prompting}
Ju, C., Han, T., Zheng, K., Zhang, Y., Xie, W.: Prompting visual-language models for efficient video understanding. In: ECCV (2022)

\bibitem{ju2023multi}
Ju, C., Li, Z., Zhao, P., Zhang, Y., Zhang, X., Tian, Q., Wang, Y., Xie, W.: Multi-modal prompting for low-shot temporal action localization. arXiv  (2023)

\bibitem{kuhn1955hungarian}
Kuhn, H.W.: The hungarian method for the assignment problem. Naval research logistics quarterly  (1955)

\bibitem{li2022align}
Li, D., Li, J., Li, H., Niebles, J.C., Hoi, S.C.: Align and prompt: Video-and-language pre-training with entity prompts. In: CVPR (2022)

\bibitem{li2024egoexo}
Li, Y.M., Huang, W.J., Wang, A.L., Zeng, L.A., Meng, J.K., Zheng, W.S.: Egoexo-fitness: Towards egocentric and exocentric full-body action understanding. ECCV  (2024)

\bibitem{li2024continual}
Li, Y.M., Zeng, L.A., Meng, J.K., Zheng, W.S.: Continual action assessment via task-consistent score-discriminative feature distribution modeling. TCSVT  (2024)

\bibitem{lin2021learning}
Lin, C., Xu, C., Luo, D., Wang, Y., Tai, Y., Wang, C., Li, J., Huang, F., Fu, Y.: Learning salient boundary feature for anchor-free temporal action localization. In: CVPR (2021)

\bibitem{lin2023univtg}
Lin, K.Q., Zhang, P., Chen, J., Pramanick, S., Gao, D., Wang, A.J., Yan, R., Shou, M.Z.: Univtg: Towards unified video-language temporal grounding. In: ICCV (2023)

\bibitem{lin2024rethinking}
Lin, K.Y., Ding, H., Zhou, J., Peng, Y.X., Zhao, Z., Loy, C.C., Zheng, W.S.: Rethinking clip-based video learners in cross-domain open-vocabulary action recognition. arXiv  (2024)

\bibitem{lin2024diversifying}
Lin, K.Y., Du, J.R., Gao, Y., Zhou, J., Zheng, W.S.: Diversifying spatial-temporal perception for video domain generalization. NeurIPS  (2024)

\bibitem{lin2024human}
Lin, K.Y., Zhou, J., Zheng, W.S.: Human-centric transformer for domain adaptive action recognition. TPAMI  (2024)

\bibitem{lin2019bmn}
Lin, T., Liu, X., Li, X., Ding, E., Wen, S.: Bmn: Boundary-matching network for temporal action proposal generation. In: ICCV (2019)

\bibitem{lin2018bsn}
Lin, T., Zhao, X., Su, H., Wang, C., Yang, M.: {BSN: Boundary Sensitive Network for Temporal Action Proposal Generation}. In: ECCV (2018)

\bibitem{lin2017focal}
Lin, T.Y., Goyal, P., Girshick, R., He, K., Doll{\'a}r, P.: Focal loss for dense object detection. In: ICCV (2017)

\bibitem{liu2022end}
Liu, X., Wang, Q., Hu, Y., Tang, X., Zhang, S., Bai, S., Bai, X.: End-to-end temporal action detection with transformer. TIP  (2022)

\bibitem{loshchilov2017decoupled}
Loshchilov, I., Hutter, F.: Decoupled weight decay regularization. ICLR  (2017)

\bibitem{luo2023towards}
Luo, D., Huang, J., Gong, S., Jin, H., Liu, Y.: Towards generalisable video moment retrieval: Visual-dynamic injection to image-text pre-training. In: CVPR (2023)

\bibitem{miech2019howto100m}
Miech, A., Zhukov, D., Alayrac, J.B., Tapaswi, M., Laptev, I., Sivic, J.: Howto100m: Learning a text-video embedding by watching hundred million narrated video clips. In: ICCV (2019)

\bibitem{moon2023query}
Moon, W., Hyun, S., Park, S., Park, D., Heo, J.P.: Query-dependent video representation for moment retrieval and highlight detection. In: CVPR (2023)

\bibitem{nag2022zero}
Nag, S., Zhu, X., Song, Y.Z., Xiang, T.: Zero-shot temporal action detection via vision-language prompting. In: ECCV (2022)

\bibitem{paszke2019pytorch}
Paszke, A., Gross, S., Massa, F., Lerer, A., Bradbury, J., Chanan, G., Killeen, T., Lin, Z., Gimelshein, N., Antiga, L., et~al.: Pytorch: An imperative style, high-performance deep learning library. NeurIPS  (2019)

\bibitem{phan2024zeetad}
Phan, T., Vo, K., Le, D., Doretto, G., Adjeroh, D., Le, N.: Zeetad: Adapting pretrained vision-language model for zero-shot end-to-end temporal action detection. In: WACV (2024)

\bibitem{radford2021learning}
Radford, A., Kim, J.W., Hallacy, C., Ramesh, A., Goh, G., Agarwal, S., Sastry, G., Askell, A., Mishkin, P., Clark, J., et~al.: Learning transferable visual models from natural language supervision. In: ICML (2021)

\bibitem{rao2022denseclip}
Rao, Y., Zhao, W., Chen, G., Tang, Y., Zhu, Z., Huang, G., Zhou, J., Lu, J.: Denseclip: Language-guided dense prediction with context-aware prompting. In: CVPR (2022)

\bibitem{shi2023tridet}
Shi, D., Zhong, Y., Cao, Q., Ma, L., Li, J., Tao, D.: Tridet: Temporal action detection with relative boundary modeling. In: CVPR (2023)

\bibitem{shi2022react}
Shi, D., Zhong, Y., Cao, Q., Zhang, J., Ma, L., Li, J., Tao, D.: React: Temporal action detection with relational queries. In: ECCV (2022)

\bibitem{sun2023hierarchical}
Sun, S., Gong, X.: Hierarchical semantic contrast for scene-aware video anomaly detection. In: CVPR (2023)

\bibitem{tan2021relaxed}
Tan, J., Tang, J., Wang, L., Wu, G.: Relaxed transformer decoders for direct action proposal generation. In: ICCV (2021)

\bibitem{vaswani2017attention}
Vaswani, A., Shazeer, N., Parmar, N., Uszkoreit, J., Jones, L., Gomez, A.N., Kaiser, {\L}., Polosukhin, I.: Attention is all you need. NeurIPS  (2017)

\bibitem{wang2023event}
Wang, A.L., Lin, K.Y., Du, J.R., Meng, J., Zheng, W.S.: Event-guided procedure planning from instructional videos with text supervision. In: ICCV (2023)

\bibitem{wu2023cap4video}
Wu, W., Luo, H., Fang, B., Wang, J., Ouyang, W.: Cap4video: What can auxiliary captions do for text-video retrieval? In: CVPR (2023)

\bibitem{xu2021videoclip}
Xu, H., Ghosh, G., Huang, P.Y., Okhonko, D., Aghajanyan, A., Metze, F., Zettlemoyer, L., Feichtenhofer, C.: Videoclip: Contrastive pre-training for zero-shot video-text understanding. arXiv  (2021)

\bibitem{xu2020g}
Xu, M., Zhao, C., Rojas, D.S., Thabet, A., Ghanem, B.: G-tad: Sub-graph localization for temporal action detection. In: CVPR (2020)

\bibitem{yuan2016temporal}
Yuan, J., Ni, B., Yang, X., Kassim, A.A.: Temporal action localization with pyramid of score distribution features. In: CVPR (2016)

\bibitem{zhang2023exploiting}
Zhang, C., Li, G., Qi, Y., Wang, S., Qing, L., Huang, Q., Yang, M.H.: Exploiting completeness and uncertainty of pseudo labels for weakly supervised video anomaly detection. In: CVPR (2023)

\bibitem{zhang2022actionformer}
Zhang, C.L., Wu, J., Li, Y.: Actionformer: Localizing moments of actions with transformers. In: ECCV. Springer (2022)

\bibitem{zhou2024actionhub}
Zhou, J., Liang, J., Lin, K.Y., Yang, J., Zheng, W.S.: Actionhub: a large-scale action video description dataset for zero-shot action recognition. arXiv  (2024)

\bibitem{zhou2021graph}
Zhou, J., Lin, K.Y., Li, H., Zheng, W.S.: Graph-based high-order relation modeling for long-term action recognition. In: CVPR (2021)

\bibitem{zhou2023twinformer}
Zhou, J., Lin, K.Y., Qiu, Y.K., Zheng, W.S.: Twinformer: Fine-to-coarse temporal modeling for long-term action recognition. TMM  (2023)

\end{thebibliography}
}

\end{document}